\documentclass[11pt]{article}

\usepackage[preprint]{acl}

\usepackage{times}
\usepackage{latexsym}

\usepackage[T1]{fontenc}

\usepackage[utf8]{inputenc}
\usepackage{newunicodechar}
\newunicodechar{‑}{-}
\newunicodechar{–}{--}
\newunicodechar{—}{---}
\newunicodechar{’}{'}
\newunicodechar{“}{``}
\newunicodechar{”}{''}

\usepackage{microtype}

\usepackage{inconsolata}

\usepackage{graphicx}
\usepackage{amsmath}
\usepackage{amssymb}
\usepackage{booktabs}
\usepackage{multirow}
\usepackage{subcaption}
\usepackage{fancyvrb}

\title{Membox: Weaving Topic Continuity into Long‑Range Memory for LLM Agents}

\author{
  Dehao Tao \\
  Tsinghua University \\
  \texttt{tdh23@mails.tsinghua.edu.cn} 
  \And
  Guoliang Ma \\
  Xinjiang University
  \And
  Yongfeng Huang \\
  Tsinghua University \\
  \And
  Minghu Jiang \\
  Tsinghua University
}

\begin{document}
\maketitle
\begin{abstract}
Long-term human--agent dialogues are organized by topic continuity: adjacent turns often develop the same goal, plan, problem, or event, while related activities may recur across distant sessions. Yet many LLM agent memory systems first decompose histories into isolated turns or fixed-size chunks, then compensate through enrichment, consolidation, or retrieval mechanisms still tied to semantic proximity or fragment-level records. This weakens temporal and causal organization and biases memory access toward semantic proximity rather than task- or topic-level continuity.
We introduce \emph{Membox}, a hierarchical memory architecture that instantiates topic continuity as an explicit organization layer for agent memory. Its \textbf{Topic Loom} incrementally organizes dialogue streams into boxes whose internal turns follow the same local topic, while its \textbf{Trace Weaver} links extracted events across boxes into macro-topic traces that recover recurring activities, goals, and factual developments across distant sessions.
On LoCoMo, Topic-Loom-only retrieval improves over the best Mem0/A-MEM retrieval-depth setting by 13.00 F1 points (53.95 vs. 40.95), and trace-expanded retrieval further raises F1 to 55.28; with GPT-4o, trace-expanded retrieval reaches 59.71 F1. Additional DialSim results show the same gain from adding cross-box traces in multi-party dialogue. These results show that local topic-continuity organization and macro-topic trace expansion improve long-range memory beyond semantic retrieval over fragmented records.
\end{abstract}

\section{Introduction}

Human memory and discourse are inherently structured around continuity. Cognitive accounts of episodic memory suggest that temporally adjacent events or interactions are bound into coherent episodes, preserving temporal order and causal relations \cite{miller1956magical,tulving1983elements,baddeley2000episodic}. Discourse theories further characterize such continuity as hierarchical: stable macro‑topics persist across interaction, while local micro‑topics drift as the conversation unfolds \cite{grosz1986attention,schiffrin1994approaches}.

For LLM agents, this continuity is not merely descriptive but functional: users often expect an agent to answer current questions by following an ongoing activity, goal, or concern across prior interactions, even when the wording changes. A question about whether one should continue running, for example, may depend jointly on earlier discussion of marathon training, recurring knee pain, and medical advice to reduce training intensity. These turns differ in surface semantics---exercise, pain, and medical advice---yet their relevance comes from belonging to the same evolving topic trajectory. Answering such questions therefore requires memory access that can retrieve these topically connected turns together, rather than treating each turn as an independent semantic match to the query.

\begin{figure}
    \centering
    \includegraphics[width=0.92\linewidth]{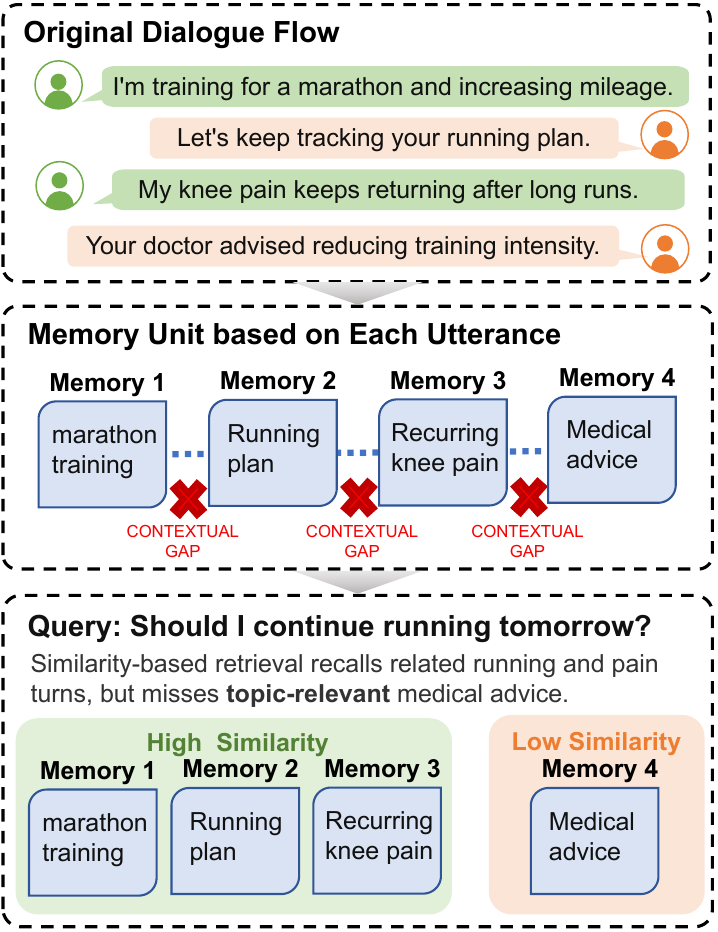}
    \caption{A representative failure of fragmentation-based memory. An ongoing running-related episode is written as separate utterance-level memories, creating contextual gaps; similarity-based retrieval recalls lexically related running and pain memories but can miss low-similarity yet topic-relevant medical advice.}
    \label{fig:existing}
\end{figure}
Most existing agent memory systems \cite{zhong2024memorybank,xu2025mem, chhikara2025mem0} do not provide this topic-level access path. Figure~\ref{fig:existing} illustrates the resulting failure in a simple utterance-level case: the running-related trajectory is written as separate memories, so similarity-based retrieval can recover lexically related running and pain memories while missing the low-similarity medical advice that is central to the later decision. This example reflects a broader \textit{fragmentation‑compensation paradigm}: dialogue continuity is first broken into separately stored fragments, and later mechanisms such as enrichment, consolidation, or retrieval attempt to compensate by adding back related context. However, because this compensation is still driven largely by semantic proximity or fragment-level operations, it can only approximate the original topic trajectory and may miss evidence whose relevance comes from continuity rather than lexical similarity. We therefore argue that long-term agent memory needs a dedicated topic-continuity organization layer between raw dialogue fragments and retrieval-time semantic matching.

This suggests a different memory-writing principle: long-term agent memory should preserve topic-continuous evidence before retrieval begins, rather than relying on retrieval-time compensation after continuity has been fragmented. By organizing dialogue at write time, temporal and causal relations that were present in the original interaction can remain available as part of the same memory structure, even when some turns are weakly similar to the later query.

To this end, \textit{Membox} introduces topic continuity as an explicit organization layer at memory construction time. Its \textbf{Topic Loom} writes micro-topic-continuous episodes online as dialogue unfolds: each incoming turn is assigned to either continue the current topic-continuous episode or open a new one before future turns are available. This converts topic-boundary detection from an offline discourse-analysis task into an online memory-writing policy, preserving locally continuous discourse before downstream retrieval begins.

\begin{figure*}
    \centering
    \includegraphics[width=1\linewidth]{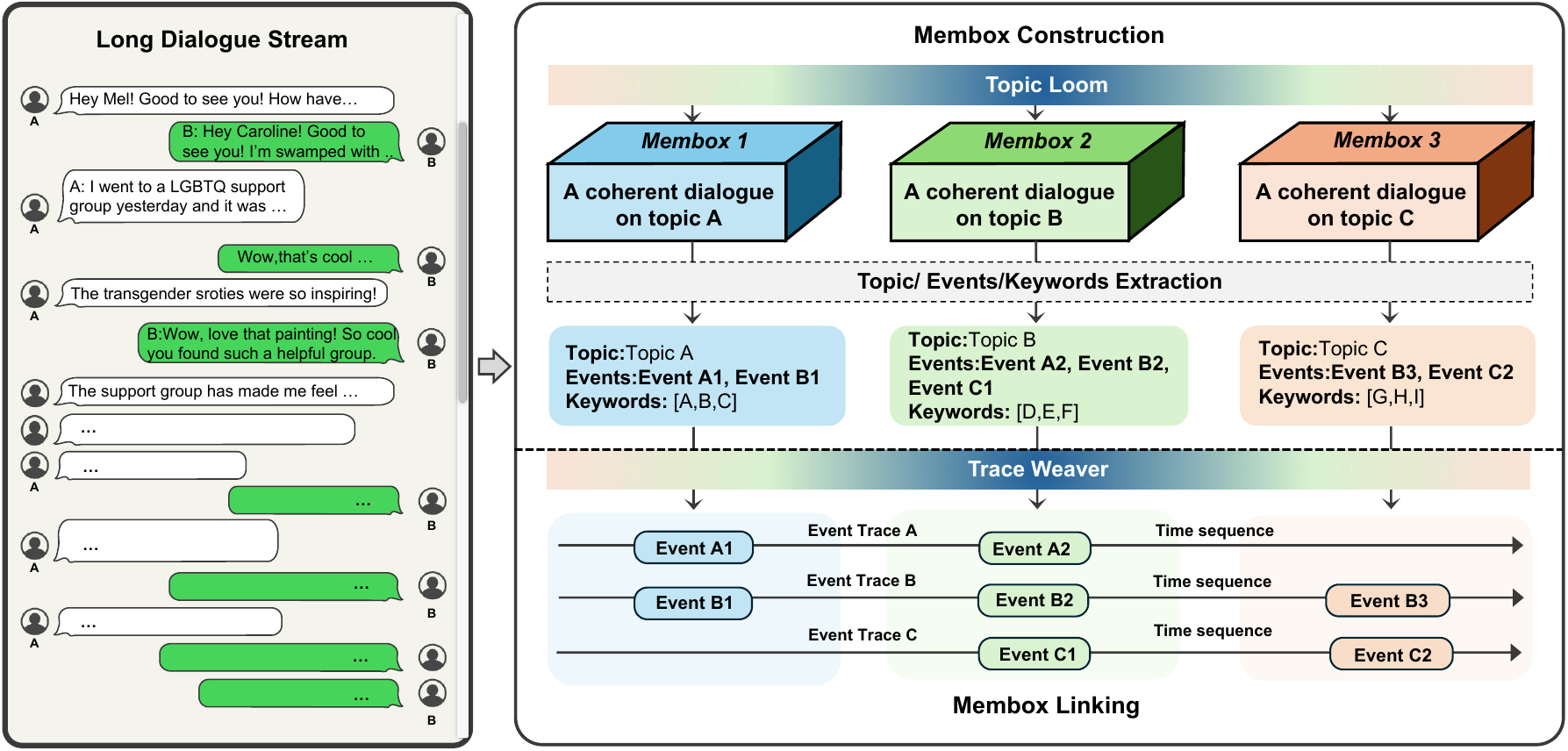}
    \caption{Overview of the Membox architecture — the Topic Loom groups micro-topic-continuous dialogue into Memboxes, while the Trace Weaver links events across Memboxes into macro-topic traces.}
    \label{fig:membox}
\end{figure*}

Membox further captures macro-topic continuity through the \textbf{Trace Weaver}, which links these episodes through their extracted events. Events act as anchors for recurring activities, goals, plans, and factual developments that may reappear after local topic shifts. The resulting architecture represents topic continuity at two coupled levels: Topic Loom preserves how a topic unfolds within a local episode, while Trace Weaver preserves how broader themes return or progress across episodes. Figure~\ref{fig:membox} illustrates this hierarchical memory architecture.

Our contributions are threefold:
\begin{itemize}
    \item We identify topic-level continuity as a missing organizational layer in agent memory and reformulate memory construction around topic-continuous dialogue episodes rather than turn- or chunk-level fragments.
    \item We propose \emph{Membox}, a two-level memory architecture whose \textbf{Topic Loom} captures micro-topic continuity within episodes, and whose \textbf{Trace Weaver} captures macro-topic continuity across episodes.
    \item We evaluate \emph{Membox} on the \emph{LoCoMo} benchmark, with additional validation on DialSim, showing that hierarchical continuity-preserving memory improves long-context dialogue QA while offering a favorable quality--context tradeoff.
\end{itemize}

\section{Related Work}
\subsection{Long-Term Memory for LLM Agents}

Long-term memory systems for LLM agents retain reusable evidence beyond the immediate context window. MemoryBank \cite{zhong2024memorybank} and Ret-LLM \cite{modarressi2023ret} use embedding-based indexing, while MemGPT \cite{packer2023memgpt} and SCM \cite{wang2023enhancing} introduce hierarchical or controller-based memory management. Recent systems emphasize adaptive updates: Mem0 \cite{chhikara2025mem0} incrementally evolves memory items, ReadAgent \cite{lee2024human} compresses long contexts into gist representations, and A-MEM \cite{xu2025mem} equips agents with decision-driven memory operations. These methods improve capacity and flexibility, but still operate over turns, chunks, summaries, or mutable records and rely on update or retrieval to compensate for fragmentation. Membox instead inserts a topic-continuity organization layer at write time before retrieval, writing conversational experience as locally coherent episodes and globally linked traces.

\subsection{Discourse Topic Segmentation}

Discourse topic segmentation places boundaries between coherent topical spans in text or dialogue. Classical methods detect boundaries through lexical cohesion and local topic shifts \cite{hearst-1997-text,galley-etal-2003-discourse}, while neural approaches model segmentation as supervised boundary prediction over fully observed sequences \cite{koshorek-etal-2018-text,arnold-etal-2019-sector,lukasik-etal-2020-text,jiang-etal-2023-superdialseg}. These settings typically assume access to the complete document or dialogue and produce a boundary label sequence.

Membox is inspired by this view of topical spans, but the agentic memory setting changes both the timing and the output of segmentation. An agent must decide online, as each turn arrives and before future context is available, whether the current message should continue the active topic-continuous episode or open a new one. The output is also not merely a boundary sequence, but an organized memory object that supports later retrieval and cross-episode trace construction.

\subsection{Retrieval-Augmented and Structured Context}

Retrieval-Augmented Generation grounds LLM outputs in external evidence retrieved at inference time \cite{lewis2020retrieval,gao2023retrieval}. Agent-like methods further let models decide when and what to retrieve through reflection, query refinement, or active retrieval policies \cite{asai2024self,jiang2023active}. Other work organizes external knowledge with structured resources such as knowledge graphs for retrieval or symbolic traversal \cite{Linders2025KG-RAG,Baek2023KAPING,Sun2024ToG,Jiang2022UniKGQA,chen2024plan}. These approaches improve access to external context, but not how conversational experience should be organized into continuity-preserving memory structures before retrieval.

\section{Method}
\subsection{Membox Construction: The Topic Loom}
\label{sec:construction}

Real-time agent systems continuously receive streams of user–agent messages, requiring an online decision about what should remain within the same organized episode. Storing each message as an isolated item is computationally simple, but it violates \textbf{topic continuity}: temporally adjacent turns that share an unfolding discourse frame may be separated before later retrieval begins.

We define a \emph{topic-continuous dialogue episode} as the local object produced by the topic-continuity organization layer: a span of consecutive turns that develop the same discourse topic, even when their surface semantics differ. We therefore use the \textbf{Topic Loom} as the construction layer of Membox: an online, LLM‑guided memory‑writing policy inspired by discourse topic segmentation. Its goal is not to infer a global topic taxonomy or cluster all semantically similar utterances, but to decide whether the incoming turn should continue the locally unfolding episode. The output is not merely a boundary decision, but a structured memory object that serves as the substrate for retrieval and later trace construction.

We maintain a small sliding window of two consecutive messages—one user utterance and one agent response—over the most recent messages in the current unsealed box, and use it for topic continuity classification.
Upon arrival of a new message \(M_{k+1}\), the Loom queries an LLM:
\[
c_{k+1} \leftarrow \mathrm{LLM}\big( \text{window},\ M_{k+1},\ P_{\mathrm{cont}} \big),
\]
where \(P_{\mathrm{cont}}\) is the classification prompt shown in Appendix Table~\ref{tab:information_extraction}, and  
\(c_{k+1} \in \{\mathrm{continuous},\ \mathrm{partial\ shift},\ \mathrm{discontinuous}\}\).  
In practice, most segment-worthy transitions are labeled as partial shifts rather than discontinuities. This is expected in continuous multi-turn conversations, where new topics often emerge through residual links to prior context rather than abrupt resets. For memory construction, partial shifts are treated as topic breaks because they indicate a noticeable change in conversational focus, even when the current turn remains locally related to previous turns.

If the label is \emph{continuous}, the message is appended to the current box.  
If it is \emph{partial shift} or \emph{discontinuous}, the current box is sealed, and a new unsealed box is created with \(M_{k+1}\) as its first entry.
To avoid degenerate single-message boxes, we require each new box to contain at least one adjacent response when available. This minimal-context rule is applied before subsequent topic-continuity classification resumes, ensuring that brief utterances are still stored with enough local dialogue context for later interpretation.

When a box transitions to the sealed state, the Loom uses the extraction prompt \(P_{\mathrm{extract}}\) (Appendix Table~\ref{tab:gg_construction_prompt}) to produce its structured representation  
\(B = \{ M,\ \mathrm{topic},\ \mathrm{events},\ \mathrm{keywords} \}\).  
The event set \(\mathrm{events}(B) = \{ e_1, e_2, \dots \}\) is extracted from the messages in \(B\) and captures concrete actions, factual developments, or plans within the box’s topic, such as marathon training, recurring knee pain, or reducing training intensity in the running example. Since our memory design centers on topic continuity, extracting events provides a natural, fine‑grained representation of each topic, while keywords supply supplementary descriptive details.
The extracted event set \(E(B)\) becomes the input to the \textbf{Trace Weaver} stage (\S\ref{sec:linking}).

\subsection{Membox Linking: The Trace Weaver}
\label{sec:linking}

The \textbf{Trace Weaver} is the macro-continuity layer of Membox. Whereas the Topic Loom binds consecutive turns into locally coherent episodes, the Trace Weaver links sealed Memboxes into cross-episode traces when their events indicate the recurrence, progression, or transformation of a broader theme.

This separation reflects two coupled dimensions of topic continuity. The Topic Loom captures micro-level continuity by preserving consecutive turns that belong to the same unfolding topic. The Trace Weaver captures macro-level continuity by organizing temporally separated Memboxes around recurring activities, goals, plans, and factual developments. Together, the two layers convert dialogue history into a hierarchy of local episodes and long-range topic traces.

Formally, after the Topic Loom seals a Membox $B_{new}$, we obtain its set of extracted events
\[
E^{(new)} = \{ e_1, e_2, \dots, e_p \}.
\]
Let $\mathcal{T} = \{ T_1, T_2, \dots, T_q \}$ denote the set of existing traces, and $E^{(T_i)}$ the events stored in trace $T_i$.

\paragraph{Trace Initialization (if $\mathcal{T}=\varnothing$).}
If there are no existing traces, we pass $E^{(new)}$ to an LLM with the initialization prompt $P_{\mathrm{init}}$ (Appendix Table~\ref{tab:choose_prompt}), clustering the events into one or more new traces:
\[
\mathcal{T} \leftarrow \mathcal{T} \cup \mathrm{LLM}\big( E^{(new)} \,\|\, P_{\mathrm{init}} \big).
\]
This establishes initial macro-topic traces for subsequent linking.

\paragraph{Event-to-Trace Voting.}
For each event \(e_k \in E^{(new)}\), we first identify the trace containing the most semantically similar stored event:
\[
T^*(e_k) = \arg\max_{T_i \in \mathcal{T}} \ \Big[ \max_{e' \in E^{(T_i)}} \ \mathrm{sim}(e_k, e') \Big],
\]
where $\mathrm{sim}(\cdot,\cdot)$ is cosine similarity in embedding space.
Each event therefore contributes one nearest trace as a vote. Taking the union of these votes yields a compact candidate trace set,
\[
\mathcal{C}(B_{new}) = \{T^*(e_k) \mid e_k \in E^{(new)}\},
\]
which limits LLM verification to traces that are locally supported by at least one event.
\paragraph{LLM Batch Verification.}
For each candidate trace $T_i \in \mathcal{C}(B_{new})$, we pass both (a) the trace's existing events $E^{(T_i)}$ and (b) the full set of events in the current box $E^{(new)}$ to the LLM with the verification prompt $P_{\mathrm{verify}}$ (Appendix Table~\ref{tab:relation_pruning}).
The LLM jointly considers topic context and event semantics to decide which events from $E^{(new)}$ should be appended to $T_i$. Thus, although each event selects only one trace during candidate generation, verification compares every new event with every trace in the candidate set; an event may be accepted into multiple traces when it fits multiple macro-topic continuities.
This batch decision process allows cross‑event reasoning within the same box, capturing cases where related events reinforce each other’s topical fit.

\paragraph{Secondary Trace Initialization.}
Events from $E^{(new)}$ not accepted into any existing traces form $E_{\mathrm{unlinked}}$.
If $E_{\mathrm{unlinked}} \neq \varnothing$, they are re‑passed to $P_{\mathrm{init}}$ to form new traces.  

In our design, traces do not form a single linear chain: an event may legitimately belong to multiple traces, reflecting the branching and intersecting nature of real discourse. Within a single Membox, different events can be assigned to distinct traces, since a local episode may contribute to several recurring macro-topics. The same architecture can support different notions of macro-level continuity, such as causal chains, temporal progressions, or role-based interaction networks, by altering the linking objective and similarity criteria.

\begin{table*}[t]
  \centering
  \caption{Main LoCoMo results. \emph{Membox-Compact} retrieves Topic Loom boxes and provides only box content to the QA model, while \emph{Membox-Trace} augments retrieved boxes with Trace Weaver events. Both modes use content top-$k=10$; Trace additionally uses event top-$k=2$. Entries marked with $^*$ (\texttt{Mem0}$^*$ and \texttt{A-MEM}$^*$) represent our local re-implementations. For these re-implemented baselines, we tune retrieval depth $k \in \{5, 10, 20, 30\}$ and report the optimal performance. The best result for each model, category, and metric is highlighted in \textbf{bold}.}
  \resizebox{\textwidth}{!}{%
  \begin{tabular}{@{}l l *{8}{c}@{}}
    \toprule
    \multirow{2}{*}{Model} &
    \multirow{2}{*}{Method} &
    \multicolumn{8}{c}{Category} \\
    \cmidrule(lr){3-10}
    & &
    \multicolumn{2}{c}{Multi-Hop} &
    \multicolumn{2}{c}{Temporal} &
    \multicolumn{2}{c}{Open Domain} &
    \multicolumn{2}{c}{Single Hop} \\
    \cmidrule(lr){3-4} \cmidrule(lr){5-6} \cmidrule(lr){7-8} \cmidrule(lr){9-10}
    & & F1 & BLEU-1 & F1 & BLEU-1 & F1 & BLEU-1 & F1 & BLEU-1 \\
    \midrule
    GPT-4o-mini & LoCoMo & 25.02 & 19.75 & 18.41 & 14.77 & 12.04 & 11.16 & 40.36 & 29.05 \\
    GPT-4o-mini & ReadAgent & 9.15 & 6.48 & 12.60 & 8.87 & 5.31 & 5.12 & 9.67 & 7.66 \\
    GPT-4o-mini & MemoryBank & 5.00 & 4.77 & 9.68 & 6.99 & 5.56 & 5.94 & 6.61 & 5.16 \\
    GPT-4o-mini & MemGPT & 26.65 & 17.72 & 25.52 & 19.44 & 9.15 & 7.44 & 41.04 & 34.34 \\
    GPT-4o-mini & A-MEM  & 27.02 & 20.09 & 45.85 & 36.67 & 12.14 & 12.00 & 44.65 & 37.06 \\
    GPT-4o-mini & A-MEM$^*$ & 27.08 & 20.46 & 29.14 & 24.08 & 16.60 & 13.80 & 40.70 & 32.63 \\
    GPT-4o-mini & Mem0 & 38.72 & 27.13 & 48.93 & 40.51 & 28.64 & 21.58 & 47.65 & 38.72 \\
    GPT-4o-mini & Mem0$^*$ & 36.83 & 26.50 & 34.52 & 26.38 & 22.57 & 16.54 & 46.89 & 37.63 \\
    GPT-4o-mini & \textbf{Membox-Compact} & 39.88 & 26.39 & 58.03 & 45.17 & 27.96 & 20.15 & 60.09 & 47.45 \\
    GPT-4o-mini & \textbf{Membox-Trace} & \textbf{41.19} & \textbf{27.49} & \textbf{59.63} & \textbf{46.52} & \textbf{30.36} & \textbf{22.52} & \textbf{61.18} & \textbf{48.99} \\
    \midrule
    GPT-4o & LoCoMo & 28.00 & 18.47 & 9.09  & 5.78  & 16.47 & 14.80 & 61.56 & \textbf{54.19} \\
    GPT-4o & ReadAgent & 14.61 & 9.95 & 4.16 & 3.19 & 8.84 & 8.37 & 12.46 & 10.29 \\
    GPT-4o & MemoryBank & 6.49 & 4.69 & 2.47 & 2.43 & 6.43 & 5.30 & 8.26 & 7.10 \\
    GPT-4o & MemGPT & 30.36 & 22.83 & 17.29 & 13.18 & 12.24 & 11.87 & 60.18 & 53.35 \\
    GPT-4o & A-MEM  & 32.86 & 23.76 & 39.41 & 31.23 & 17.10 & 15.84 & 48.43 & 42.97 \\
    GPT-4o & Mem0$^*$ & 42.57 & 30.92 & 44.55 & 32.60 & 23.04 & 17.62 & 48.49 & 37.00 \\
    GPT-4o & A-MEM$^*$ & 31.66 & 23.34 & 41.11 & 34.72 & 17.45 & 15.58 & 47.04 & 41.02 \\
    GPT-4o & \textbf{Membox-Compact} & 48.35 & 35.10 & 65.06 & \textbf{54.81} & 30.61 & 22.58 & 61.69 & 49.36 \\
    GPT-4o & \textbf{Membox-Trace} & \textbf{50.48} & \textbf{38.17} & \textbf{66.61} & 54.15 & \textbf{38.77} & \textbf{28.19} & \textbf{62.56} & 48.95 \\
    \bottomrule
  \end{tabular}
  }
  \label{tab:model_performance}
\end{table*}

\subsection{Retrieval}
\label{sec:retrieval}

Retrieval follows the hierarchical organization of Membox. Given a query $q$, we first rank sealed Memboxes by comparing $q$ with each box representation $R(B)=\{M,\mathrm{topic},E(B),\mathrm{keywords}\}$:
\[
\mathcal{B}_q = \operatorname{TopK}_{B \in \mathcal{B}}^{k_b}\ \mathrm{sim}\big(q, R(B)\big),
\]
where $\mathcal{B}_q$ denotes the retrieved boxes and provides the local episodic evidence for the query.
This use of similarity differs from retrieval over fragmented memories: similarity is used only to access already organized memory structures. Once a box is selected, the QA model receives the whole topic-continuous episode, including turns that may be weakly similar to the query but are necessary for interpreting the episode.

When trace information is used, retrieval is refined through the event layer of $\mathcal{B}_q$. Let
\[
\mathcal{E}_q = \bigcup_{B \in \mathcal{B}_q} E(B)
\]
be the candidate event set induced by the retrieved boxes. We select the most query-relevant events,
\[
\mathcal{E}^{\mathrm{top}}_q = \operatorname{TopK}_{e \in \mathcal{E}_q}^{k_e}\ \mathrm{sim}(q,e),
\]
and retrieve the traces associated with these events:
\[
\mathcal{T}_q = \{T_i \in \mathcal{T} \mid E^{(T_i)} \cap \mathcal{E}^{\mathrm{top}}_q \neq \varnothing\}.
\]
Thus, trace retrieval is a structured expansion of the box-level results: selected boxes determine the candidate events, and the most query-relevant events determine the long-range traces used as additional evidence. In Trace mode, the QA context consists of the retrieved box contents plus all events contained in the selected traces; raw dialogue from non-retrieved boxes is not added.
Similarity therefore serves as an entry point into the topic-continuity organization layer, rather than as a substitute for that organization.

\begin{table*}[h]
\centering
\caption{
Memory base statistics.  
Utterances: total number of utterances;  
Tok Ratio: (constructed memory tokens) / (original dialogue tokens);  
MB\#: Membox count;  
Utter/MB: utterances per Membox;  
Tok/MB: text tokens per Membox.  
Note: “token” here refers to text length, not LLM processing tokens.  
Tokens are segmented simply by spaces in this analysis.
}
\begin{tabular}{lrrrrr}
\toprule
Method & Utterances & Tok Ratio & MB\# & Utter/MB & Tok/MB \\
\midrule
Mem0   w/ GPT-4o-mini & 5882 & 1.19 & -    & -     & - \\
Mem0   w/ GPT-4o      & 5882 & 1.20 & -    & -     & - \\
A-MEM   w/ GPT-4o-mini & 5882 & 1.72 & -    & -     & - \\
A-MEM   w/ GPT-4o      & 5882 & 1.73 & -    & -     & - \\
Membox w/ GPT-4o-mini & 5882 & 1.24 & 892  & 6.59 & 342.98 \\
Membox w/ GPT-4o      & 5882 & 1.19 & 1206 & 4.88 & 252.63 \\
\bottomrule
\end{tabular}
\label{tab:memory}
\end{table*}

\section{Experiment}

\subsection{Dataset and Evaluation}
We use the \textit{LoCoMo} benchmark \cite{maharana2024evaluating} as our primary evaluation platform. \textit{LoCoMo} presents a significant long-context challenge, featuring dialogues that average 35 sessions and approximately 9,000 tokens. Such scale requires robust long-range retrieval and stable reasoning across extended sequences.
Following the benchmark protocol, we evaluate four dimensions:
\textbf{Single-hop Retrieval}: extracting specific facts from one session.
\textbf{Multi-hop Reasoning}: synthesizing information across multiple sessions.
\textbf{Temporal Reasoning}: understanding event sequences and durations within the dialogue flow.
\textbf{Open-domain QA}: integrating dialogue history with external commonsense knowledge. The original dataset includes adversarial unanswerable questions, but these specify answers that should not be produced and lack gold answers, so EM/F1 cannot be computed under the standard generation protocol. Treating ``unanswerable'' as an answer would require a separate abstention protocol and break comparability; we therefore exclude this category from the main benchmark.

We compare \textbf{Membox} with six competitive baselines: \textbf{LoCoMo} \cite{maharana2024evaluating}, \textbf{ReadAgent} \cite{lee2024human}, \textbf{MemoryBank} \cite{zhong2024memorybank}, \textbf{MemGPT} \cite{packer2023memgpt}, \textbf{A-MEM} \cite{xu2025mem}, and \textbf{Mem0} \cite{chhikara2025mem0}. Appendix~\ref{sec:dialsim} further evaluates cross-dataset generalization on DialSim \cite{kim2024dialsim}. We report \textbf{F1} for precision--recall balance in answer generation and \textbf{BLEU-1} for lexical overlap with references.

\subsection{Implementation Details}
We utilize \textbf{text-embedding-3-small} for text embedding and OpenAI's \textbf{GPT-4o} and \textbf{GPT-4o-mini} as the backbone LLMs across all experiments.

For a fair comparison, we locally deploy and evaluate A-MEM \cite{xu2025mem} and Mem0 \cite{chhikara2025mem0}, test retrieval depths $k \in \{5, 10, 20, 30\}$, and report their best scores. For Membox, the main comparison uses fixed content top-$k=10$ and two inference modes: \emph{Membox-Compact}, which retrieves Topic Loom boxes as evidence, and \emph{Membox-Trace}, which further expands the context with Trace Weaver events from related macro-topic traces using event top-$k=2$. Retrieval-depth sensitivity is reported in Table~\ref{tab:topk}, and the full trace-expansion sweep in Appendix Table~\ref{tab:mode_sensitivity_full}. The evaluation follows two phases: 1) Memory Construction with each system's default prompts; 2) QA \& Inference with the same backbone LLM.

\subsection{Empirical Results}

Table~\ref{tab:model_performance} reports the main LoCoMo comparison. Membox-Compact isolates the effect of Topic Loom: by retrieving locally topic-continuous boxes, it already outperforms strong baselines on most categories across both GPT-4o-mini and GPT-4o. This comparison directly tests whether adding a topic-continuity organization layer before retrieval provides better evidence than increasing retrieval depth over fragmented memory records.

Membox-Trace adds the macro-continuity layer by expanding retrieved boxes with Trace Weaver events. This mode further improves F1 across all four categories for both backbones, with the largest gains on open-domain QA and consistent gains elsewhere. BLEU-1 generally follows the same trend, although GPT-4o shows small tradeoffs on temporal and single-hop BLEU-1. These results indicate that Trace Weaver is not merely a temporal-question add-on; it provides useful cross-episode context when the answer depends on recurring activities, goals, or factual developments.

Overall, the main comparison supports the hierarchical design: Topic Loom provides a compact local evidence substrate, while Trace Weaver offers an optional macro-topic expansion for higher-recall answering. Appendix~\ref{sec:dialsim} reports an additional DialSim evaluation on multi-speaker dialogue, where the same Compact/Trace pattern yields higher F1 than session retrieval and full-dialogue-context baselines while using far fewer context tokens.

\begin{figure*}[t]
    \centering
    \includegraphics[width=1\textwidth]{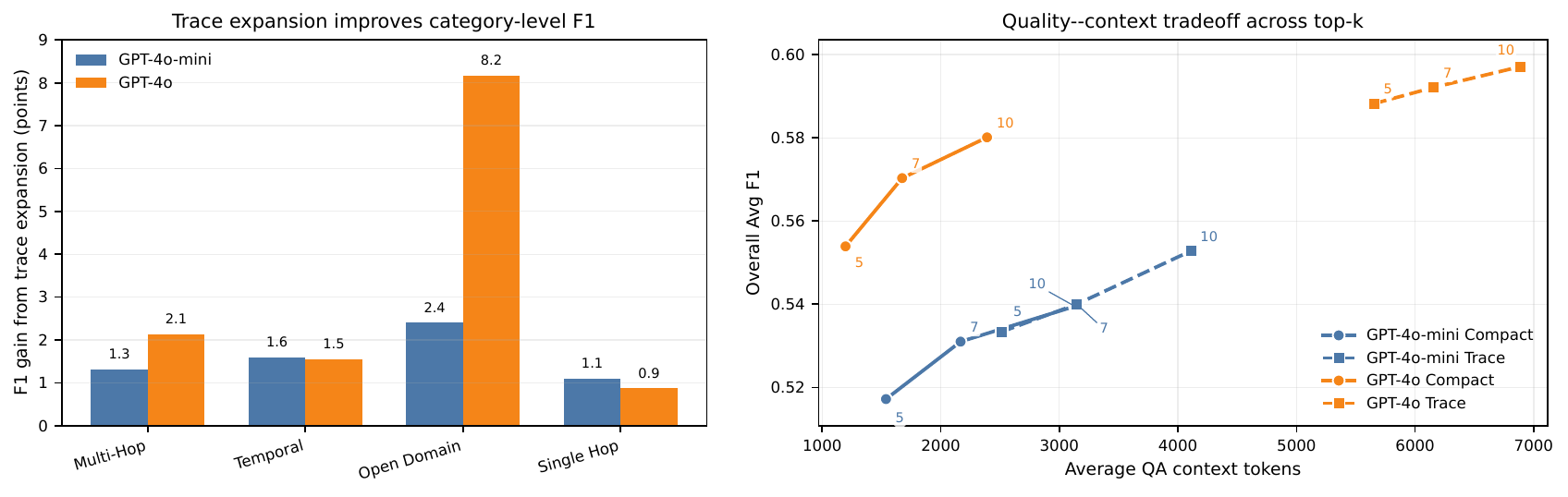}
    \caption{Effect of Trace Weaver at inference time. Left: F1 gains from adding trace expansion to compact box retrieval under content top-$k=10$ and event top-$k=2$. Right: overall quality--context tradeoff across content top-$k \in \{5,7,10\}$; numbers indicate content top-$k$, and Trace uses event top-$k=2$.}
    \label{fig:mode_tradeoff}
\end{figure*}

\subsection{Retrieval-Depth Analysis}
Table~\ref{tab:topk} compares retrieval-depth sensitivity on GPT-4o-mini. Membox-Compact improves steadily as more Topic Loom boxes are retrieved, but most of the gain appears within the first few boxes: F1 increases from 0.3988 at top-$k=1$ to 0.4941 at top-$k=3$, then grows more gradually to 0.5395 at top-$k=10$. Under comparable QA context budgets, Membox also achieves higher generation quality than Mem0 and A-MEM, suggesting that topic-continuous boxes provide denser evidence than isolated memory records.

\begin{table*}[t]
\centering
\caption{Overall retrieval-depth comparison on GPT-4o-mini. Membox rows use \emph{Membox-Compact}. All rows are evaluated on 1540 QA instances.}
\begin{tabular}{lrrrr}
\toprule
Method & top-$k$ & Avg F1 & Avg BLEU-1 & Avg ctx tok \\
\midrule
Mem0 & 5  & 0.3836 & 0.2970 & 331.14  \\
Mem0 & 10 & 0.3986 & 0.3102 & 656.89  \\
Mem0 & 20 & 0.4035 & 0.3155 & 1306.11  \\
Mem0 & 30 & 0.4095 & 0.3193 & 1955.00 \\
A-MEM & 5  & 0.3063 & 0.2524 & 1238.77  \\
A-MEM & 10 & 0.3277 & 0.2926 & 2449.88  \\
A-MEM & 20 & 0.3365 & 0.3273 & 4873.67  \\
A-MEM & 30 & 0.3441 & 0.3488 & 7246.66 \\
Membox-Compact & 1 & 0.3988 & 0.3113 & 310.69  \\
Membox-Compact & 3 & 0.4941 & 0.3818 & 917.03  \\
Membox-Compact & 5 & 0.5172 & 0.3970 & 1538.10 \\
Membox-Compact & 7 & 0.5310 & 0.4070 & 2166.88 \\
Membox-Compact & 10 & 0.5395 & 0.4142 & 3130.72 \\
\bottomrule
\end{tabular}
\label{tab:topk}
\end{table*}

\subsection{Analysis on Memory Construction}
\paragraph{Memory Size Analysis}
As shown in Table~\ref{tab:memory}, the final memory size produced by \textbf{Membox} is comparable to \textbf{Mem0} and notably smaller than \textbf{A-MEM}, while organizing dialogue into topic-continuous boxes. Each box contains about \textbf{4--6} utterances on average, reducing context fragmentation without substantially increasing the memory footprint. Since Mem0 and A-MEM do not construct box units, MB-level statistics are reported only for Membox.

\paragraph{LLM Consumption Analysis}
Appendix Tables~\ref{tab:memory_llm} and~\ref{tab:trace_llm} report LLM token consumption during construction and linking. Membox performs lightweight online continuity checks as messages arrive, but structured extraction and memory writing occur at the box level rather than as full per-utterance updates. This explains why Membox consumes fewer LLM tokens per utterance than Mem0 and A-MEM while preserving local dialogue context. The two backbones produce different box partitions, but both maintain the same efficiency pattern.

\subsection{Analysis on Membox Linking}
Tables~\ref{tab:memory_llm} and~\ref{tab:trace_llm} separate Topic Loom construction from Trace Weaver linking. Linking is more expensive because it verifies event-to-trace assignments against existing global traces, costing about twice the construction stage per box, yet the combined preprocessing cost per utterance remains below the reported Mem0 and A-MEM costs in our implementation.

Fig.~\ref{fig:mode_tradeoff} summarizes the inference-time tradeoff introduced by Trace Weaver. Trace expansion improves overall F1 from 0.5395 to 0.5528 on GPT-4o-mini and from 0.5801 to 0.5971 on GPT-4o. The gain comes with additional context tokens, so Membox can be used either as a compact Topic Loom retriever or as a trace-expanded retriever when higher-recall macro-topic evidence is useful. Detailed mode statistics and the full trace-expansion sweep are provided in Appendix Tables~\ref{tab:mode_tradeoff} and~\ref{tab:mode_sensitivity_full}.

\section{Conclusions}
This paper addresses the challenge of topic continuity in human–agent dialogue—the tendency for adjacent turns to form coherent thematic episodes. Existing agent memory systems follow a fragmentation–compensation paradigm that first decomposes dialogue into isolated turns or fixed-size chunks and then compensates through similarity-based enrichment or retrieval, resulting in structural discontinuities and biases toward surface‑level similarity rather than topic continuity.
We propose Membox, a hierarchical memory architecture that augments agent memory with a topic-continuity organization layer. The Topic Loom organizes incoming dialogue into locally coherent episodes through sliding-window monitoring, while the Trace Weaver links sealed episodes across discontinuities to recover recurring macro-topics and long-range event timelines.
Experiments on LoCoMo show that Membox improves F1 across the evaluated QA categories over strong baselines such as Mem0 and A‑MEM, while BLEU-1 generally follows the same trend with small metric-specific tradeoffs in a few settings. Compact retrieval demonstrates the value of topic-continuous boxes, while trace-expanded retrieval further improves long-range QA by adding macro-topic evidence. These results demonstrate that modeling topic continuity as a memory-construction principle yields more coherent, efficient, and temporally grounded LLM agents.

\section*{Limitations}

Membox represents macro-topic continuity primarily through event-based traces. This design is effective for recurring activities, plans, factual developments, and other long-range dependencies that can be naturally expressed as events. However, real long-term agent memory may also require continuity along other dimensions, such as user preferences, interpersonal relations, affective states, stable personality traits, or evolving constraints that are not always event-like. Our current Trace Weaver therefore should be viewed as one instantiation of the broader topic-continuity organization layer rather than a complete representation of all forms of long-term memory. Extending Membox to support multiple trace types is an important direction for future work.


\bibliography{custom}

\appendix

\section{Appendix}
\label{sec:appendix}

\subsection{LLM Usage Statement}
The large language model was used solely for grammar checks and polishing, and no other purposes.

\subsection{Additional DialSim Evaluation}
\label{sec:dialsim}

To test whether the observed gains transfer beyond LoCoMo, we additionally evaluate Membox on the Friends subset of DialSim \cite{kim2024dialsim}, a multi-speaker casual-dialogue benchmark. We use dialogues from Seasons 1--5, covering 118 episodes. For each episode, we sample 5 easy and 5 hard questions, yielding 1025 questions after duplicate removal. QA uses GPT-4o-mini with embedding-based retrieval. To avoid prompt tuning to the new benchmark, we reuse the LoCoMo memory-construction prompts; the answering prompt follows the DialSim paper. DialSim also includes unanswerable cases with gold answer sets, so they can be evaluated with standard F1.

\begin{table}[!htbp]
    \centering
    \caption{Additional DialSim results on Friends. Membox rows use top-10 retrieval. Baseline results are from the DialSim paper; context tokens for Session Retrieval are not reported there.}
    \begin{tabular}{lrr}
    \toprule
    Method & Avg F1 & Avg ctx tok \\
    \midrule
    Session Retrieval & 45.47 & -- \\
    Full Dialogue Context & 48.11 & $\sim$128k \\
    Membox-Compact & 52.78 & 826.53 \\
    Membox-Trace & \textbf{55.33} & 1872.54 \\
    \bottomrule
    \end{tabular}
    \label{tab:dialsim}
\end{table}

As shown in Table~\ref{tab:dialsim}, Membox-Compact outperforms the DialSim baselines while using 826.53 average context tokens, compared with approximately 128k tokens for full-dialogue context. Trace expansion further improves F1 to 55.33 with 1872.54 average context tokens. Since DialSim contains multi-speaker casual conversations and the prompts are not tuned for it, these results suggest that hierarchical topic-continuity memory is not specific to LoCoMo.

\subsection{Preprocessing Cost Details}

Tables~\ref{tab:memory_llm} and~\ref{tab:trace_llm} report detailed LLM usage during memory construction and trace linking. They complement the main analysis by separating the online Topic Loom construction cost from the Trace Weaver linking cost.

\begin{table}[h]
\centering
\caption{
LLM call statistics during memory base construction.  
MB\#: Membox count;  
Tok/MB: LLM tokens consumed per Membox;  
Tok/Ut: LLM tokens consumed per utterance.  
}
\resizebox{\columnwidth}{!}{%
\begin{tabular}{@{}lrrr@{}}
\toprule
Method & MB\# & Tok/MB & Tok/Ut \\
\midrule
Mem0 w/ GPT-4o-mini & - & - & 2115.85 \\
Mem0 w/ GPT-4o      & - & - & 1923.17 \\
A-MEM w/ GPT-4o-mini & - & - & 1755.57 \\
A-MEM w/ GPT-4o      & - & - & 1526.39 \\
Membox w/ GPT-4o-mini & 892  & 1557.44 & 236.18 \\
Membox w/ GPT-4o      & 1206 & 1241.61 & 254.57 \\
\bottomrule
\end{tabular}
}

\label{tab:memory_llm}
\end{table}

\begin{table}[h]
\centering
\caption{
LLM usage statistics for Membox linking.  
MB\#: Membox count;  
Calls/MB: LLM calls per Membox;  
Tok/MB: tokens consumed per Membox.
}
\begin{tabular}{@{}lrrr@{}}
\toprule
Model & MB\# & Calls/MB & Tok/MB  \\
\midrule
GPT-4o-mini & 892  & 2.30 & 3133.56  \\
GPT-4o      & 1206 & 0.88 & 2716.89 \\
\bottomrule
\end{tabular}

\label{tab:trace_llm}
\end{table}

\begin{table}[h]
\centering
\caption{Overall quality--context tradeoff for the two Membox inference modes. Both modes use content top-$k=10$; Trace additionally uses event top-$k=2$. Ctx denotes average QA context tokens.}
\resizebox{\columnwidth}{!}{%
\begin{tabular}{@{}llrrr@{}}
\toprule
Model & Mode & F1 & BLEU-1 & Ctx \\
\midrule
GPT-4o-mini & Compact & 0.5395 & 0.4142 & 3130.72 \\
GPT-4o-mini & Trace & 0.5528 & 0.4289 & 4108.85 \\
GPT-4o & Compact & 0.5801 & 0.4621 & 2390.16 \\
GPT-4o & Trace & 0.5971 & 0.4676 & 6887.84 \\
\bottomrule
\end{tabular}
}
\label{tab:mode_tradeoff}
\end{table}

\subsection{Trace-Expansion Sensitivity}

Table~\ref{tab:mode_sensitivity_full} reports the full sensitivity sweep for trace-expanded retrieval. Across both GPT-4o and GPT-4o-mini, adding Trace Weaver evidence consistently improves over compact box retrieval at the same content top-$k$, confirming that macro-topic traces provide useful long-range evidence beyond the retrieved local episodes. Increasing the event top-$k$ generally raises F1 but also increases context length, so the main experiments use event top-$k=2$ as a balanced setting: it captures most of the trace benefit without adding all linked events to the QA context.

\begin{table*}[htbp]
\centering
\caption{Full overall sensitivity of Compact and Trace inference modes. Trace rows use the same content retrieval as Compact and vary the event top-$k$. All rows are evaluated on 1540 QA instances.}
\scriptsize
\begin{tabular}{llrrrrr}
\toprule
Model & Mode & Content top-$k$ & Event top-$k$ & Avg F1 & Avg BLEU-1 & Avg ctx tok \\
\midrule
GPT-4o & Compact & 5 & -- & 0.5539 & 0.4394 & 1196.79 \\
GPT-4o & Compact & 7 & -- & 0.5703 & 0.4505 & 1674.21 \\
GPT-4o & Compact & 10 & -- & 0.5801 & 0.4621 & 2390.16 \\
GPT-4o & Trace & 5 & 1 & 0.5826 & 0.4616 & 5362.73 \\
GPT-4o & Trace & 7 & 1 & 0.5881 & 0.4617 & 5827.23 \\
GPT-4o & Trace & 10 & 1 & 0.5908 & 0.4643 & 6555.07 \\
GPT-4o & Trace & 5 & 2 & 0.5882 & 0.4664 & 5652.71 \\
GPT-4o & Trace & 7 & 2 & 0.5921 & 0.4649 & 6155.77 \\
GPT-4o & Trace & 10 & 2 & 0.5971 & 0.4676 & 6887.84 \\
GPT-4o & Trace & 5 & all & 0.5960 & 0.4711 & 5928.89 \\
GPT-4o & Trace & 7 & all & 0.5919 & 0.4646 & 6459.29 \\
GPT-4o & Trace & 10 & all & 0.5986 & 0.4697 & 7225.96 \\
\midrule
GPT-4o-mini & Compact & 5 & -- & 0.5172 & 0.3970 & 1538.10 \\
GPT-4o-mini & Compact & 7 & -- & 0.5310 & 0.4070 & 2166.88 \\
GPT-4o-mini & Compact & 10 & -- & 0.5395 & 0.4142 & 3130.72 \\
GPT-4o-mini & Trace & 5 & 1 & 0.5301 & 0.4164 & 2268.74 \\
GPT-4o-mini & Trace & 7 & 1 & 0.5369 & 0.4187 & 2900.56 \\
GPT-4o-mini & Trace & 10 & 1 & 0.5463 & 0.4253 & 3858.10 \\
GPT-4o-mini & Trace & 5 & 2 & 0.5334 & 0.4185 & 2515.48 \\
GPT-4o-mini & Trace & 7 & 2 & 0.5400 & 0.4217 & 3146.83 \\
GPT-4o-mini & Trace & 10 & 2 & 0.5528 & 0.4289 & 4108.85 \\
GPT-4o-mini & Trace & 5 & all & 0.5400 & 0.4193 & 4042.98 \\
GPT-4o-mini & Trace & 7 & all & 0.5447 & 0.4209 & 5065.46 \\
GPT-4o-mini & Trace & 10 & all & 0.5560 & 0.4295 & 6527.00 \\
\bottomrule
\end{tabular}
\label{tab:mode_sensitivity_full}
\end{table*}

\subsection{Category-Level Retrieval-Depth Sensitivity}
\label{sec:category_topk}

\begin{figure*}[t]
    \centering
    \begin{subfigure}[b]{0.45\textwidth}
        \centering
        \includegraphics[width=\textwidth]{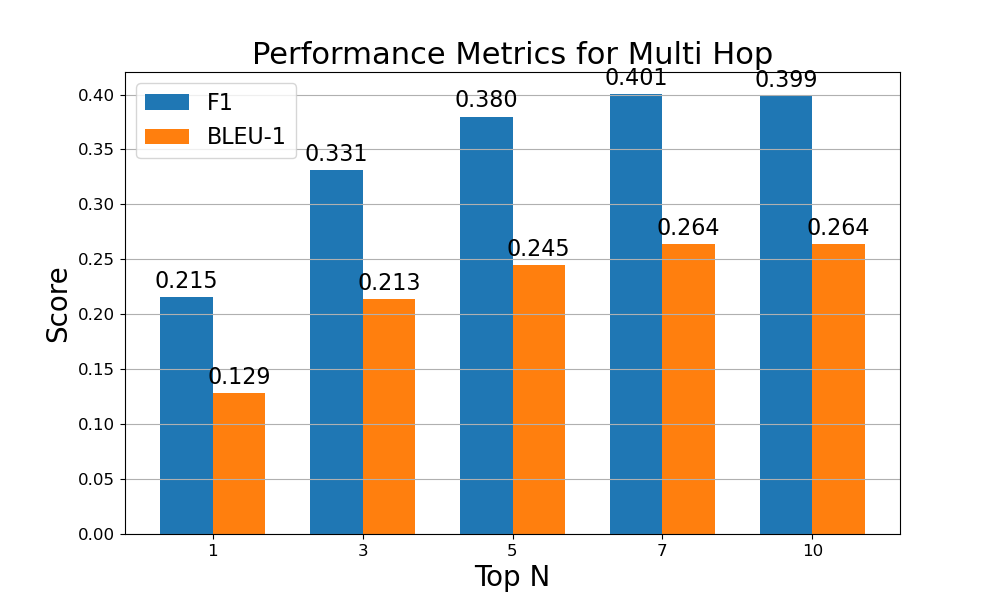}
        \caption{Multi-hop}
    \end{subfigure}
    \hfill
    \begin{subfigure}[b]{0.45\textwidth}
        \centering
        \includegraphics[width=\textwidth]{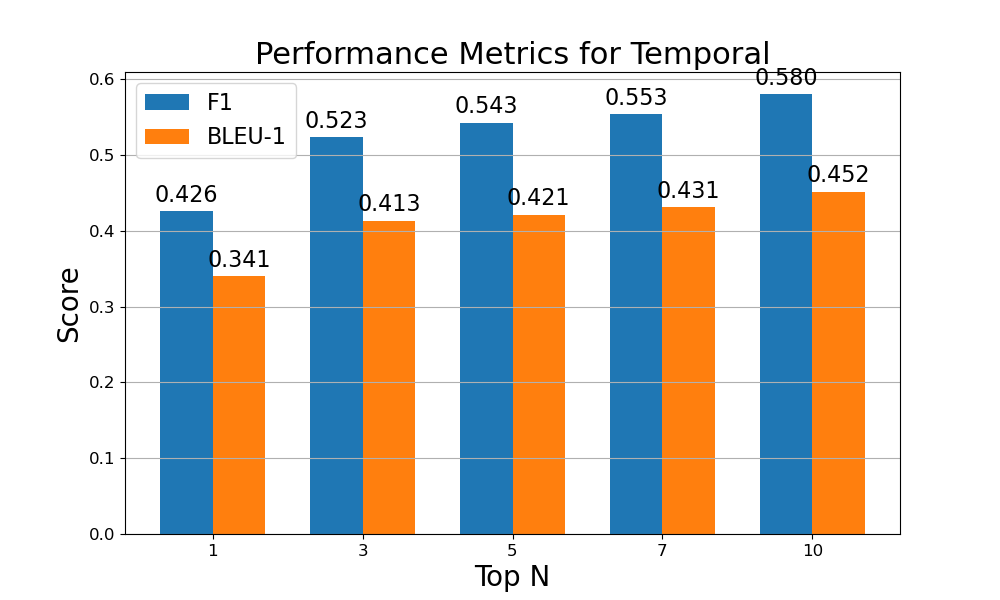}
        \caption{Temporal}
    \end{subfigure}

    \begin{subfigure}[b]{0.45\textwidth}
        \centering
        \includegraphics[width=\textwidth]{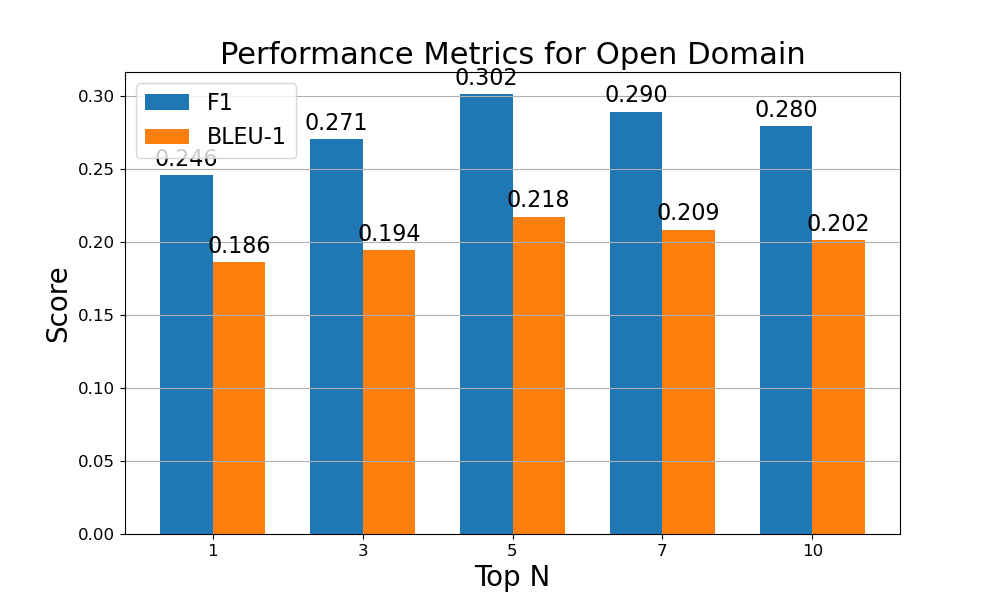}
        \caption{Open-domain}
    \end{subfigure}
    \hfill
    \begin{subfigure}[b]{0.45\textwidth}
        \centering
        \includegraphics[width=\textwidth]{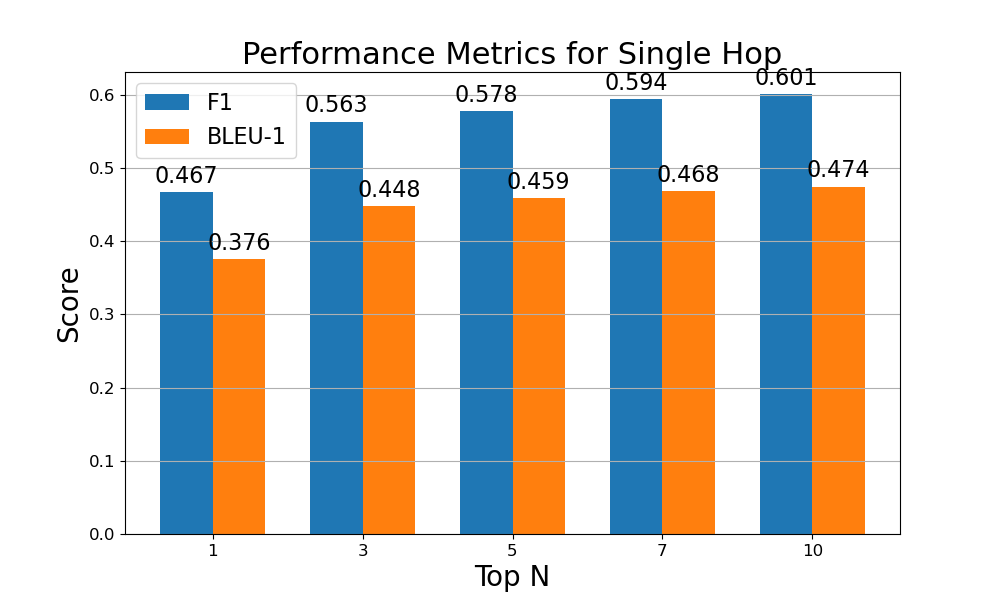}
        \caption{Single-hop}
    \end{subfigure}
    \caption{Category-level retrieval-depth sensitivity for Membox-Compact with GPT-4o-mini on LoCoMo. Top-$N$ denotes the number of retrieved Topic Loom boxes.}
    \label{fig:category_topk}
\end{figure*}

Figure~\ref{fig:category_topk} provides a category-level view of retrieval-depth sensitivity for Membox-Compact with GPT-4o-mini. The plots vary the number of retrieved Topic Loom boxes while keeping the retrieval mode fixed to compact box retrieval, so they isolate how much local topic-continuous evidence is needed for each question type.

Temporal and single-hop questions benefit steadily from larger retrieval depth, suggesting that additional topic-continuous boxes provide useful supporting evidence for event order, duration, and factual lookup. Multi-hop questions improve sharply from top-1 to top-7 and then saturate, indicating that a small set of coherent episodes already captures most of the required evidence. Open-domain questions peak around top-5 and then slightly decline, which suggests that adding more dialogue context can introduce distractors when the answer also depends on external commonsense knowledge. Overall, the category-level trends support the main retrieval-depth analysis: retrieving a few topic-continuous boxes yields most of the gain, while larger top-$k$ values mainly trade additional evidence for more context.

\subsection{Prompt Templates}

In this study, four types of prompts are employed, each serving a distinct function. PROMPT\_MSG\_CONTINUATION (Table~\ref{tab:information_extraction}) is used during memory construction to determine whether the current dialogue is continuous with the previous context, thereby deciding whether the active topic-continuous episode should be extended or sealed. PROMPT\_DIALOG\_EXTRACT (Table~\ref{tab:gg_construction_prompt}) extracts key information from the dialogue and converts it into a structured format stored in the memory module. PROMPT\_TRACE\_EVENT\_FILTER (Table~\ref{tab:relation_pruning}) and PROMPT\_TRACE\_INIT (Table~\ref{tab:choose_prompt}) are used to construct event traces.

\begin{table*}[htbp]
    \centering
    \begin{tabular}{p{0.95\linewidth}}
    \toprule
    \textbf{PROMPT MSG CONTINUATION} \\
    \midrule
    \small
    Please determine whether the current message continues with the main topic of the previous messages. Only answer Yes/No/Partially Shifted.\\[0.5em]
    
    \textbf{previous messages}: \texttt{ref}\\[0.5em]
    
    \textbf{current message}: \texttt{curr}\\[0.5em]
    
    \textbf{Answer:}\\
    \bottomrule
        \end{tabular}
        \caption{PROMPT MSG CONTINUATION}
    \label{tab:information_extraction}
\end{table*}

\begin{table*}[htbp]
    \centering

    \begin{minipage}{0.95\linewidth}
    \small
    \begin{tabular}{@{}p{\linewidth}@{}}
    \toprule
    \textbf{PROMPT DIALOG EXTRACT} \\
    \midrule
    Please analyze the relationships between the following entities in the given sentence.\\[3pt]
    
    Generate a structured analysis of the provided dialog by performing the following tasks:\\[3pt]
    
    1. \textbf{Identifying salient keywords}: Extract 3-8 most salient nouns, named entities, and key terminology that represent core concepts. Avoid common words (e.g., ``good'', ``see'') and prioritize specificity.\\[3pt]
    
    2. \textbf{Determining the core topic}: In one clear phrase, state the primary subject or objective of the discussion based on the actual content.\\[3pt]
    
    3. \textbf{Extracting explicit event and plan mentions}: Identify and list only the \textbf{events, factual developments, or specific future plans} that are \textbf{explicitly mentioned} in the dialog. Follow these strict rules:\\[3pt]
    
    \hspace{1em}3.1. \textbf{Focus on Verbatim or Near-Verbatim Content}: Each extracted item must be directly grounded in the dialog text. Do not infer, summarize, or combine information to create new ``events.''\\[3pt]
    
    \hspace{1em}3.2. \textbf{Distinguish Event Types}:\\
    \hspace{2em}- \textbf{Past/Completed Events}: Actions or occurrences that are stated as having happened (e.g., ``I went to...'', ``We completed the project'').\\
    \hspace{2em}- \textbf{Established Facts/Changes}: Concrete facts or changes presented as already true (e.g., ``I am now the team lead'', ``The system is down'').\\
    \hspace{2em}- \textbf{Explicit Future Plans}: Specific plans for the future mentioned by the speakers (e.g., ``We will meet on Friday'', ``I'm planning to visit Paris'').\\[3pt]
    
    \hspace{1em}3.3. \textbf{Exclude Non-Events}: Do NOT include:\\
    \hspace{2em}- General states of being (e.g., ``I'm swamped'', ``I'm happy'').\\
    \hspace{2em}- Questions, greetings, or expressions of intent without a plan (e.g., ``We should talk sometime'').\\
    \hspace{2em}- Vague aspirations or possibilities.\\[3pt]
    
    \hspace{1em}3.4. \textbf{Framing}: Phrase each extracted item as a concise, standalone clause that captures the core of what was mentioned.\\[3pt]
    
    \textbf{Output Format}: Provide the analysis as a valid JSON object with the following exact keys:\\[3pt]
    
    \begin{verbatim}
{
  "keywords": [
    "keyword1", 
    "keyword2", 
    ...
  ],
  "topic": "clear topic phrase",
  "explicit_mentions": [
    "A mentioned past event or established fact",
    "A mentioned specific future plan"
  ]
}
    \end{verbatim}
    
    Content to analyze: \{\texttt{text}\}\\
    \bottomrule
    \end{tabular}
        \caption{PROMPT DIALOG EXTRACT}
            \label{tab:gg_construction_prompt}
    \end{minipage}
\end{table*}

\begin{table*}[htbp]
    \centering
    \scriptsize  
    \begin{tabular}{@{}p{0.95\linewidth}@{}}
    \toprule
    \textbf{PROMPT\_TRACE\_EVENT\_FILTER} \\
    \midrule
    You are a narrative coherence analyzer for constructing and maintaining event memory chains. Your task is to filter events from a new event list (Event List B) that are directly related to an existing event chain (Event Chain A).\\[0.3em]
    
    \textbf{Core Task:}\\
    Event Chain A represents an existing sequence of events (could be one or multiple events). Event List B is a set of newly observed events. Analyze each event in B to determine whether it should:\\
    1. Serve as a \textbf{direct continuation} of Event Chain A (directly related to A's core narrative)\\
    2. Be considered \textbf{unrelated} to Event Chain A (independent or belonging to a different event stream)\\[0.3em]
    
    \textbf{Analysis Principles:}\\
    - Identify the \textbf{core theme/activity} from Event Chain A's overall narrative\\
    - Assess narrative continuity: Does the event from B advance, develop, or resolve A's core activity?\\
    - Consider temporal/causal logic: Does the event naturally follow A's chain in time or logic?\\[0.3em]
    
    \textbf{Decision Criteria:}\\
    An event from B is \textbf{related} to Event Chain A if it:\\
    1. Continues the \textbf{same core activity} as A's chain (not just similar topic)\\
    2. Provides \textbf{progress, outcome, solution, or direct consequence} to A's chain\\
    3. Is a \textbf{logical/temporal successor} to A's chain\\[0.3em]
    
    An event from B is \textbf{unrelated} to Event Chain A if it:\\
    1. Initiates a \textbf{new, distinct activity} (even if topic is similar)\\
    2. Is a \textbf{parallel but independent} event to A's core activity\\
    3. Concerns a \textbf{different aspect} unrelated to A's main thread\\
    4. Is a \textbf{generic response} without specific progression\\[0.3em]
    
    \textbf{Output Format:}\\
    Strictly use this JSON format:\\
    
    \begin{minipage}{\linewidth}
    \begin{verbatim}
{
    "chain_summary": "Brief summary of Event Chain A's core theme (1-2 sentences)",
    "related_events": ["Exact text of related events from B"],
    "unrelated_events": ["Exact text of unrelated events from B"],
    "reasoning": {
        "related_reasons": ["Brief explanation for each related event"],
        "unrelated_reasons": ["Brief explanation for each unrelated event"]
    }
}
    \end{verbatim}
    \end{minipage}\\[0.3em]

    \textbf{Example 1:}\\
    Event Chain A: ["I'm planning a weekend hike", "I checked the weather forecast", "I bought hiking shoes"]\\
    Event List B: ["I mapped out the hiking route", "I replied to work emails", "I contacted hiking partners", "Went to see a movie in the evening"]\\[0.3em]
    
    Output:\\
    
    \begin{minipage}{\linewidth}
    \begin{verbatim}
{
    "chain_summary": "Preparations for a weekend hiking trip",
    "related_events": ["I mapped out the hiking route", "I contacted hiking partners"],
    "unrelated_events": ["I replied to work emails", "Went to see a movie in the evening"],
    "reasoning": {
        "related_reasons": [
            "Mapping the route is a concrete step in hike preparation",
            "Contacting partners directly advances the hiking activity"
        ],
        "unrelated_reasons": [
            "Work emails concern a different domain (work vs. recreation)",
            "Movie watching is a separate leisure activity"
        ]
    }
}
    \end{verbatim}
    \end{minipage}\\[0.3em]
    
    \textbf{Example 2:}\\
    Event Chain A: ["The project encountered technical difficulties", "The team met to discuss solutions"]\\
    Event List B: ["I researched relevant documentation", "Decided to adopt a new framework", "Had pizza for lunch", "Client sent new requirements"]\\[0.3em]
    
    Output:\\
    
    \begin{minipage}{\linewidth}
    \begin{verbatim}
{
    "chain_summary": "Addressing technical challenges in a project",
    "related_events": ["I researched relevant documentation", "Decided to adopt a new framework"],
    "unrelated_events": ["Had pizza for lunch", "Client sent new requirements"],
    "reasoning": {
        "related_reasons": [
            "Researching documentation directly addresses the technical problem",
            "Deciding on a new framework represents a solution to the technical challenge"
        ],
        "unrelated_reasons": [
            "Lunch is a routine activity unrelated to problem-solving",
            "New client requirements initiate a separate work thread"
        ]
    }
}
    \end{verbatim}
    \end{minipage}\\[0.3em]

    \textbf{Now analyze:}\\
    Event Chain A: \{\texttt{content\_a}\} (Note: This is an existing event chain)\\
    Event List B: \{\texttt{content\_b}\} (Note: This is a new event list)\\
    Output your analysis.\\
    \bottomrule
    \end{tabular}
    \caption{PROMPT\_TRACE\_EVENT\_FILTER}
    \label{tab:relation_pruning}
\end{table*}

\begin{table*}[htbp]
    \centering
    \begin{tabular}{@{}p{0.95\linewidth}@{}}
    \toprule
    \textbf{PROMPT TRACE INIT} \\
    \midrule
    \small
    You are an event chain constructor for building coherent memory structures. Your task is to analyze a set of events and organize them into logical chains.\\[0.5em]
    
    \textbf{Task:}\\
    Given a set of events, identify the primary narrative thread and any associated events that form a coherent event chain.\\[0.5em]
    
    \textbf{Process:}\\
    1. Analyze all events to identify the most prominent theme or activity\\
    2. Connect events that share temporal, causal, or thematic relationships\\
    3. Form the most coherent sequence possible\\
    4. Identify any events that don't fit into the main narrative thread\\[0.5em]
    
    \textbf{Output Format:}\\
    
    \begin{minipage}{\linewidth}
    \begin{Verbatim}[fontsize=\scriptsize]
{
    "primary_chain": ["Events forming the most coherent narrative, in logical order"],
    "secondary_chains": [["Other potential chains, if any"]],
    "isolated_events": ["Events that don't fit into any chain"],
    "chain_summary": "Brief description of the primary chain's theme and context"
}
    \end{Verbatim}
    \end{minipage}\\[0.5em]
    
    \textbf{Examples:}\\[0.5em]
    
    \textbf{Example 1:}\\
    Events: ["I woke up at 7 AM", "I checked my email", "I had breakfast", "Then I went for a run"]\\[0.5em]
    
    Output:\\
    
    \begin{minipage}{\linewidth}
    \begin{Verbatim}[fontsize=\scriptsize]
{
    "primary_chain": ["I woke up at 7 AM", "I had breakfast", "Then I went for a run"],
    "secondary_chains": [],
    "isolated_events": ["I checked my email"],
    "chain_summary": "Morning routine including waking, eating, and exercise"
}
    \end{Verbatim}
    \end{minipage}\\[0.5em]
    
    \textbf{Example 2:}\\
    Events: ["Started a new project at work", "Researched design patterns", "Met with the client", "Created initial wireframes", "Had lunch with a colleague"]\\[0.5em]
    
    Output:\\
    
    \begin{minipage}{\linewidth}
    \begin{Verbatim}[fontsize=\scriptsize]
{
    "primary_chain": ["Started a new project at work",
    "Researched design patterns", "Created initial wireframes"],
    "secondary_chains": [["Met with the client"]],
    "isolated_events": ["Had lunch with a colleague"],
    "chain_summary": "Work project initiation and initial design phase"
}
    \end{Verbatim}
    \end{minipage}\\[0.5em]
    
    \textbf{Now analyze:}\\
    Events: \{\texttt{events}\}\\
    Output your analysis in JSON format.\\[2em]
    
    \bottomrule
    \end{tabular}
    \caption{PROMPT TRACE INIT}
    \label{tab:choose_prompt}
\end{table*}

\end{document}